\documentclass[runningheads]{llncs}
\usepackage[T1]{fontenc}
\usepackage{graphicx}
\usepackage{booktabs}
\usepackage[misc]{ifsym}
\newcommand{\corr}{(\Letter)}
\usepackage{mwe}
\usepackage{lineno}
\usepackage{hyperref}
\usepackage{inconsolata}
\usepackage{amssymb}
\usepackage{amsfonts}
\usepackage{multirow}
\usepackage{pifont}
\usepackage{microtype}
\usepackage{times}
\usepackage{latexsym}
\usepackage[utf8]{inputenc}
\usepackage{caption}
\usepackage{xcolor}
\usepackage{subcaption}
\usepackage{wrapfig}
\DeclareCaptionFormat{cont}{#1 (cont.)#2#3\par}
\newcommand{\cmark}{\ding{51}}%
\newcommand{\xmark}{\ding{55}}%

\makeatletter
\newcommand*{\inlineequation}[2][]{%
  \begingroup
    \refstepcounter{equation}%
    \ifx\\#1\\%
    \else
      \label{#1}%
    \fi
    \relpenalty=10000 %
    \binoppenalty=10000 %
    \ensuremath{%
      #2%
    }%
    ~\@eqnnum
  \endgroup
}
\makeatother

\title{Task Prompt Vectors: Effective Initialization through Multi-Task Soft Prompt Transfer}

\begin{document}

\titlerunning{Task Prompt Vectors}

\author{Robert Belanec\inst{1,2} \corr \and
Simon Ostermann\inst{3} \and
Ivan Srba\inst{2} \and 
Maria Bielikova\inst{2}}

\authorrunning{Belanec et al.}

\institute{Faculty of Information Technology, Brno University of Technology, Brno, Czechia
\and
Kempelen Institute of Intelligent Technologies, Bratislava, Slovakia \email{\texttt{\{name.surname\}}@kinit.sk}
\and
German Research Center for Artificial Intelligence (DFKI), Saarbrücken, Germany \email{\texttt{simon.ostermann@dfki.de}}}

\tocauthor{Robert Belanec, Simon Ostermann, Ivan Srba, Maria Bielikova}
\toctitle{Task Prompt Vectors: Effective Initialization through Multi-Task Soft Prompt Transfer}

\maketitle              % typeset the header of the contribution

\begin{abstract}
Prompt tuning is a parameter-efficient method for adapting large language models (LLMs), where only a small continuous soft prompt is finetuned. In recent works, soft prompts have usually been trained in a task-specific way, leaving their multi-task capabilities underexplored. Our work aims to make soft prompts more \emph{task modular} based on recent research on task vectors, where arithmetic operations are applied on full model weights to achieve the desired multi-task performance. To this end, we introduce \emph{Task Prompt Vectors}, created by the element-wise difference between weights of tuned soft prompts and their random initialization.
Experimental results on an extensive set of 19 datasets show that task prompt vectors can be used in low-resource settings to initialize prompt tuning on similar tasks effectively. In addition, we show that task prompt vectors are independent of the random initialization of prompt tuning on 3 different language model architectures. This key property of random initialization independence allows \emph{prompt arithmetics} with the pre-trained vectors from different tasks. In this way, the arithmetic addition of task prompt vectors from multiple tasks represents a competitive and computationally more effective alternative to state-of-the-art solutions.
\end{abstract}

\section{Introduction}
Standard fine-tuning methods change the weights of a pre-trained language model (PLM) to increase its performance on a downstream task. There is a strong trend of improving model performance by increasing the number of parameters, which leads to a steep increase in computational resources required for training (e.g., GPT-3 \cite{brown2020language} having 175 billion parameters). Besides this, large language models also require significant amounts of training data, which especially benefits high-resource languages \cite{costa2022no}.

To address the problem of the increasing number of parameters, \textit{Parameter-Efficient Fine-Tuning (PEFT)} methods \cite{lester-etal-2021-power,houlsby2019parameter,hu2022lora} were introduced, capable of solving multiple problems even with small amounts of labeled data while training only a fraction of the model parameters (e.g., for RoBERTa base \cite{liu2019roberta}, prompt tuning \cite{lester-etal-2021-power} is training only 0.5\% parameters, and LoRA \cite{hu2022lora} is training only 0.7\% of parameters \cite{xu2023parameter}). The key concept that makes many PEFT methods effective is their \textit{task modularity} -- single modules can be trained for diverse tasks and then just be swapped out inside of the same model.

\begin{figure*}
    \begin{centering}
        \includegraphics[width=0.7\textwidth]{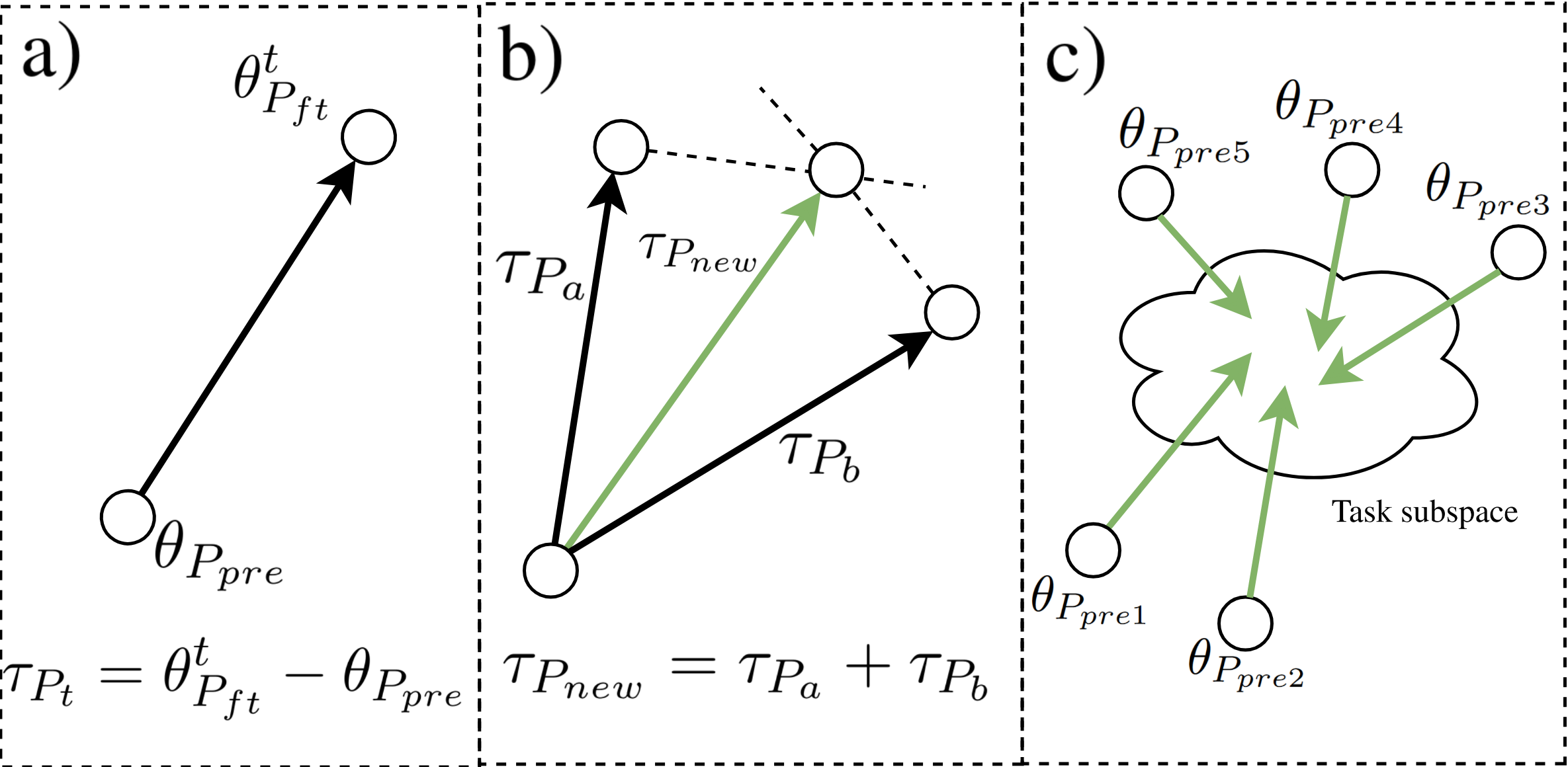}
        \caption{An illustration of task prompt vector creation and the combination via addition that we include in our work. (a) A task prompt vector is created by subtracting the soft prompt initialization weights $\theta_{P_{pre}}$ from the soft prompt weights after prompt tuning $\theta^t_{P_{ft}}$ (Section \ref{sec:method}, eq. \ref{eq:tpv}). (b) A combination via the addition of two task prompt vectors $\tau_{P_a}$ and $\tau_{P_b}$ resulting in $\tau_{P_{new}}$ (Section \ref{sec:method}, eq. \ref{eq:tpv_comb}). (c) Different task prompt vectors point into the same subspace in the embedding space of PLM (Section \ref{sec:exp_create}). The circles represent different random initializations. \label{fig:first_page}}
    \end{centering}
\end{figure*}

Some of the recent PEFT methods \cite{xu2023parameter,lester-etal-2021-power,asai-etal-2022-attempt}  are based on fine-tuning \textit{soft prompts}. Soft prompts are trainable (parametrized) weights that are prepended to the input embeddings while training the model. \textit{Prompt tuning} is one of the most widely used variants of soft prompt-based tuning of large language models (LLMs).

In contrast to other PEFT methods, most soft prompt-based methods lack sufficient multi-task modularity, requiring the training process to be fully or partially repeated for each newly added task \cite{vu-etal-2022-spot,wang2023multitask}. Other PEFT methods, while keeping their relatively high modularity, usually lack robustness, and their performance depends on the quality and the number of pre-trained soft prompts \cite{asai-etal-2022-attempt}. Moreover, creating a soft prompt for multiple tasks often decreases the overall multi-task performance and requires further fine-tuning. 

We build on findings of research on \textit{task vector arithmetics} \cite{ilharco2022editing}, a suite of methods for efficiently modifying the behavior of pre-trained LLMs through \textbf{task vectors}. A task vector represents a direction in the LLM's activation space, which is obtained by subtracting the weights of a base model from its fine-tuned version. Moving along in this direction enhances performance on the corresponding task. Task vector arithmetic has mostly been applied to the full weights of computer vision models and older NLP models like T5 \cite{raffel2020exploring} and GPT-2 \cite{radford2019language} trained from the same initialization, with only a limited set of machine unlearning experiments.

In our work, we fully extend this idea to the NLP domain, concretely to the efficient and modular techniques of prompt tuning \cite{lester-etal-2021-power}, and propose the novel concept of \textbf{task prompt vectors}. Task prompt vectors are created by subtracting soft prompt initializations from their fine-tuned versions, enabling \textit{prompt arithmetics} on top of the task prompt vectors (see Figure \ref{fig:first_page}). We thoroughly investigate the properties of task prompt vectors and demonstrate their functionality in combining pairs of task prompt vectors while evaluating their in-distribution and out-of-distribution performance in full and limited data scenarios.

Our main contributions and findings are\footnote{To support the replicability of our work, we provide a repository to store all of our implementation, results, and supplementary material: \href{https://github.com/kinit-sk/task-prompt-vectors}{https://github.com/kinit-sk/task-prompt-vectors}}:
\begin{itemize}
    \item We introduce the novel concept of \textbf{task prompt vectors} created from fine-tuned soft prompts as a method of weight interpolation that leverages findings from task vectors. In addition, we investigate vector arithmetics on such task prompt vectors based on simple arithmetic operations as a method to reinforce PLMs to solve multi-task problems. 
    \item We provide a comprehensive investigation of task prompt vector properties on 17 natural language understanding (NLU) and 2 natural language generation (NLG) datasets separated into 8 task types and demonstrate important properties of task prompt vectors. We show that their random initialization independence makes them robust and universally applicable, while their similarity across related problems provides a necessary base for efficient cross-task transfer.
    \item We show that task prompt vectors allow efficient prompt tuning initializations by leveraging multi-task combinations of the pre-trained task prompt vectors using the task prompt vector arithmetics. Experimental results show that, especially in zero- or few-shot settings, task-prompt-vector-based initializations perform better or at par with closely related techniques like SPoT (Soft Prompt Transfer learning \cite{vu-etal-2022-spot}) for specific tasks while achieving high multi-task modularity.
\end{itemize}

\section{Related Work}
\label{sec:related_work}

\paragraph{Soft prompt-based fine-tuning.} After the introduction of prompt tuning \cite{lester-etal-2021-power} and prefix tuning \cite{li-liang-2021-prefix} many new soft prompt-based methods \cite{gu-etal-2022-ppt,liu2023gpt,shi2024dept} were introduced. Some of these methods focus on task knowledge transfer (e.g., SPoT \cite{vu-etal-2022-spot} or cross-model transfer \cite{su-etal-2022-transferability}) and task combinations (e.g., ATTEMPT \cite{asai-etal-2022-attempt}, MPT \cite{wang2023multitask}, or BMTPT \cite{lee-etal-2023-bayesian}). These can be classified as works on PEFT weight interpolations to increase the performance of prompt tuning in single or multi-task settings. However, they do not represent the tasks as vectors in the embedding space and require further training of the added parameters.

\paragraph{Model weights interpolation.}
Model weight interpolation \cite{frankle2020linear,wortsman2022robust} is a widely discussed topic in the literature since it enables combining knowledge of different fine-tuned models without or with a small amount of training. Authors of tasks vectors \cite{ilharco2022editing} show that it is possible to combine multiple task vectors created from fine-tuned models and still maintain the overall multi-task performance. Work \cite{ortiz2024task} focuses mostly on improving task vectors by showing that training models in their tangent space contributes to the weight disentanglement and increases the performance of full model task arithmetic. Another subcategory for weight interpolation can be model merging \cite{stoica_zipit_2024,matena_merging_2022,li_branch-train-merge_2022,davari_model_2023}. In the work \cite{rame2023model}, the authors propose a strategy of merging multiple model weights from pre-trained sets of auxiliary tasks as initialization to multiple parallel fine-tunings to enhance out-of-distribution generalization. Most of these works on model weights interpolation usually focus only on the weights of the whole model or particular weights (e.g., classification heads, activation layers) of the pre-trained model.

There are also works on weight interpolation of PEFT methods \cite{zhang2023composing,chronopoulou2023language,pfeiffer-etal-2021-adapterfusion,10.1109/TASLP.2024.3430545}, but not many of them focus on interpolation using task vectors. In the work \cite{klimaszewski2024no}, the authors present a way of combining pre-trained adapters using task vector arithmetics, but the method lacks the investigation of the dependency of their method on the random initialization of adapters. Therefore, it may require training of specific adapters from the same random initialization, which significantly limits their re-use potential.

To the best of our knowledge, there is no research on task vectors in the context of soft prompt-based fine-tuning. In this work, we address this drawback by building on the existing knowledge on prompt tuning and task vectors.

\section{Task Prompt Vectors}
\label{sec:method}

\paragraph{Background.} Prompt tuning, as introduced in \cite{lester-etal-2021-power}, casts tasks as text generation, modeling a probability $Pr(Y|X)$, where $X$ is a sequence of input tokens and $Y$ is a sequence of output tokens (for classifications tasks, e.g., representing the class label). The generation $Pr_\theta(Y|X)$ is parametrized by the model weights $\theta$. Prompting adds extra information to the generation process by prepending a series of tokens (prompt) $P$ to the input $X$, such that the model maximizes the probability of getting current $Y$ in $Pr_\theta(Y|[P;X])$, while keeping the parameters $\theta$ frozen. Prompt tuning adds another parameter $\theta_P$ to the equation, which parametrizes the prompt. During the training, only $\theta_P$ is typically updated as a negative log-likelihood loss is optimized as:

\begin{equation}
\label{eq:pt}
\mathcal{L}_{PT} = - \sum_i log Pr_{\theta,\theta_P}(Y_i|[P;X_i]) %function is optimized.
\end{equation}

As a method of adapting model weights without training, task vectors \cite{ilharco2022editing} were proposed. A task vector is defined as the element-wise difference between the pre-trained weights and the weights after fine-tuning a complete model. Task vectors can then be applied to any model weights $\theta$ of the same dimensionality (architecture) by element-wise addition. The representation of task vectors in the weight space of the model has the same properties as standard vectors. Therefore, it is possible to include them in arithmetic expressions like addition, negation, or combinations via the addition of two or more vectors. We build on findings from \cite{ilharco2022editing} and \cite{lester-etal-2021-power} in the following sections. 

\paragraph{Task prompt vector definition.} Let $T_1, ..., T_t$ be a set of source tasks and $\theta_{P_1}, ..., \theta_{P_i}$be a set of random soft prompt weights initializations. Intuitively, the random soft prompt weights initializations are random points in the embedding space of the PLM. During prompt tuning, we move each of these points into a task sub-space, such that the objective function in equation \ref{eq:pt} is minimized. This is repeated for each task $t \in T$. These points are further denoted as \textit{task prompts} -- soft prompts fine-tuned by prompt tuning to a set of downstream tasks. We define the straight trajectory from the initial random point to the task prompt as our \textit{task prompt vector} (see part a) of Figure \ref{fig:first_page}).

Let $\theta_{P_{pre}} \in \mathbb{R}^d$ be the weights of the soft prompt randomly initialized from the embedding vocabulary of a PLM, and $\theta^t_{P_{ft}} \in \mathbb{R}^d$ be the weights of the soft prompt $P$ fine-tuned on a specific task $t$, using the standard prompt tuning formula. We formulate the task prompt vector $\tau_{P_t}$ for soft prompt $P$ and task as an element-wise difference:
\begin{equation}
    \label{eq:tpv}
    \tau_{P_t} = \theta^t_{P_{ft}} - \theta_{P_{pre}}
\end{equation}
Applying a task prompt vector to the soft prompt weights of equal size would follow: 
\begin{equation}
    \label{eq:tpv_apply}
    \theta_{P_{new}} = \theta_P + \lambda\tau_{P_t},
\end{equation}
where the rescaling term $\lambda$ is a number from the interval $0 < \lambda \le 1$ and when $\lambda = 1$, then $\theta_{P_{new}} = \theta_P + \tau_{P_t} = \theta^t_{P_{ft}}$

\paragraph{Vector arithmetics with task prompt vectors.} Task prompt vectors for different tasks can be combined by simple vector addition, combining knowledge from different tasks. When we experiment with combinations, we refer to the arithmetic addition of two task prompt vectors (see part b) of Figure \ref{fig:first_page}): 
\begin{equation}
    \label{eq:tpv_comb}
    \tau_{P_{new}} = \tau_{P_a} + \tau_{P_b} 
\end{equation}

This approach clearly results in efficient task adaptation as we perform no further training but only use vector addition. Task prompt vector combinations can also be used to initialize a new task that is sufficiently similar to an already trained task. We investigate and discuss these use cases for task prompt vectors in the upcoming sections.
 
\section{Experiments}

\subsection{Experimental Setup and Implementation Details} We investigate the properties of task prompt vectors using a \textbf{T5-base} \cite{raffel2020exploring} model for all of our experiments since it is a widely used model in many PEFT related works, and it has a reasonable size to exdend experiments to a larger scale. To support the generalizability of our results, for origin dependency experiments in Section \ref{sec:exp_create}, we also include \textbf{LLaMa-3.1-8B-Instruct} \cite{dubey2024llama} and \textbf{DeepSeek-LLM-7b-chat} \cite{bi2024deepseek} models, representing two additional LLM families. Our work covers 6 types of classification problems, as well as 2 types of generation problems covered by 19 corresponding datasets, namely \textbf{natural language inference (NLI)} -- \textit{MNLI} \cite{williams-etal-2018-broad}, \textit{QNLI} \cite{wang-etal-2018-glue}, \textit{SciTail} \cite{Khot2018SciTaiLAT}, \textit{SNLI} \cite{snli:emnlp2015}, \textit{RTE} \cite{wang-etal-2018-glue}; \textbf{topic classification} -- \textit{DBPedia} \cite{dbpedia}, \textit{TREC} \cite{li-roth-2002-learning,hovy-etal-2001-toward}, \textit{AG News}, \textit{Yahoo Answers} \cite{zhang2015character}; \textbf{sentiment classification} -- \textit{SST2} \cite{socher2013recursive}, \textit{Yelp Polarity}, \textit{SST5}, \textit{IMDB} \cite{maas-etal-2011-learning}; \textbf{paraphrase classification} -- \textit{QQP}\footnote{\href{https://quoradata.quora.com/First-Quora-Dataset-Release-Question-Pairs}{https://quoradata.quora.com/First-Quora-Dataset-Release-Question-Pairs}}, \textit{MRPC} \cite{dolan2005automatically}; \textbf{grammatical correctness} -- \textit{CoLA} \cite{warstadt-etal-2019-neural}; \textbf{semantic textual similarity} -- \textit{STS-B} \cite{cer-etal-2017-semeval}; \textbf{question answering} -- \textit{SQuADv2} \cite{rajpurkar-etal-2018-know}, and \textbf{math problems solving} -- \textit{MATH} \cite{lighteval}.

For all datasets, we report macro F1 scores, with the exception of 
STS-B (evaluated by Pearson Correlation) and MATH (evaluated by RougeL score). The cosine similarity between vectors (task prompts or task prompt vectors) is measured using the average pooled weights of each vector. We average all of our results across 3 different runs (i.e., different random initializations of soft prompts). To determine the statistical significance of our results, we perform a two-sample Student's t-test \cite{student1908probable} with Bonferroni correction \cite{dunn1959confidence}. We denote the statistical significance by marking the corresponding result with an asterisk '*'. The subscript in our tables represents the standard deviation.

For the few-shot experiments (simulating limited labeled data scenarios), we randomly sub-sample from the data for the respective number of shots while keeping the class distribution. We consider \textit{shot} and \textit{sample} to be equivalent (i.e., for a 5-shot setting, we choose 5 samples overall, not 5 samples per class). When combining task prompt vectors, we evaluate their performance on the individual source tasks that formed the task combination and find the best rescaling factor $\lambda$ via held-out validation sets (i.e., we randomly sample a validation subset from the evaluation dataset and select the best performing $\lambda \in \{0.1,0.2,...,0.9,1\}$).

We provide information about ethical considerations and an impact statement in Supplementary Material A. In addition, a more detailed description of our experimental setup can be found in Supplementary Material B.

\subsection{Investigating Task Prompt Vectors Properties}
\label{sec:exp_create}
In this section, we aim to address the following research question (RQ): 

\textbf{RQ1:} \textbf{How universally can we apply task prompt vectors to a) different prompt initializations and b) different tasks?} 

There are two fundamental properties that are crucial for the effectiveness of task prompt vectors: 1) If prompt vectors should be applied universally, they must be independent of random initialization (since soft prompts are usually initialized randomly, unlike PLM for task prompts in \cite{ilharco2022editing}). 2) The similarity of task prompt vectors between similar tasks should be high enough in order to be able to combine task prompt vectors.

\begin{figure}[!t]
    \begin{subfigure}[t]{0.49\textwidth}
        \begin{centering}
            \includegraphics[width=\textwidth]{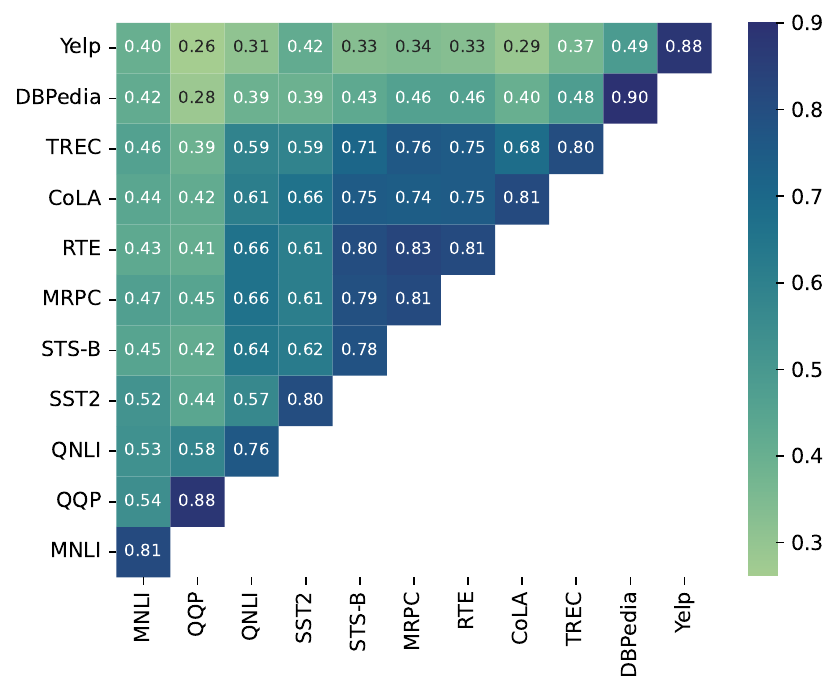}
            \caption{Cosine similarities of \textit{task prompts}. 
            \label{fig:rq1_heatmap_tp}}
        \end{centering}
    \end{subfigure}
    ~
    \begin{subfigure}[t]{0.49\textwidth}
        \begin{centering}
            \includegraphics[width=\textwidth]{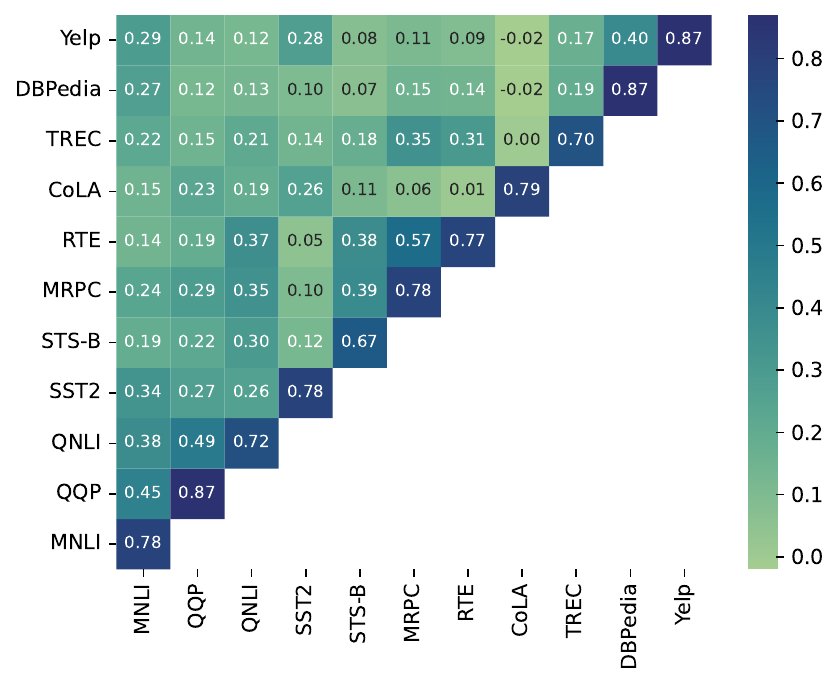}
            \caption{Cosine similarities of \textit{task prompt vectors}.
            \label{fig:rq1_heatmap_tpv}}
        \end{centering}
    \end{subfigure}
    \caption{Comparison of average cosine similarities between task prompts and task prompt vectors fine-tuned on different tasks for the T5-base model. The average is calculated across all combinations of 3 random initializations (i.e., row QNLI column MNLI was calculated as the average of all cosine similarities between MNLI and QNLI for all initialization combinations, omitting the combinations of the same vectors). The diagonal represents the cosine similarities within the same task. It provides an estimate of natural in-task variation of task prompts and task prompt vectors, against which other similarities should be compared.}
\end{figure}

To evaluate these properties, we train a set of soft prompts on specified source tasks for {inference classification} (\textit{MNLI, QNLI, RTE}), {topic classification} (\textit{DBPedia, TREC}), {sentiment classification} (\textit{SST2, Yelp Polarity}), {paraphrase classification} (\textit{QQP, MRPC}), {grammatical correctness} (\textit{CoLA}), {semantic textual similarity} (\textit{STS-B}), {question answering} (\textit{SQuADv2}) and {math problems solving} (\textit{MATH}) resulting in a set of {13 soft prompts} that were trained from a single random initialization. We sample \textit{3 random initializations} from which we create the task prompt vectors as described in equation \ref{eq:tpv}. Since SQuADv2 and MATH are more complex tasks that T5 struggles with, we report for these tasks only results for LLaMa-3.1-8B-Instruct and DeepSeek-LLM-7B-Chat. We aggregate results by averaging across random initializations in Table \ref{tab:cross_origin} and Figures \ref{fig:rq1_heatmap_tp}, \ref{fig:rq1_heatmap_tpv}. 
At first, we evaluate task prompt vectors' independence of the random initialization and continue with experiments to confirm whether task prompt vectors trained for the same task are always pointing in a similar direction of the PLM embedding space, similar to part c) of Figure \ref{fig:first_page}.

\paragraph{The performance of task prompt vectors is independent of the random initialization for the majority of observed tasks.} We conduct experiments to evaluate the performance of applying task prompt vectors to different (mixed) random initializations. For each task and each random initialization, we apply the task prompt vector (according to the equation \ref{eq:tpv_apply}) to all of the other random initializations and evaluate performance for each task prompt vector-initialization pair on the test set of the particular dataset. The aggregated results in the "Mixed init" rows in Table \ref{tab:cross_origin} differ only slightly in most tasks for all three models, compared to the results of prompt tuning in the "Original init" rows. This indicates that task prompt vectors perform well, irrespective of their initialization. The only exception is the TREC task, where the performance decreases significantly for the T5-base model. We suspect that this may be caused by the task being harder for the T5-base model to learn, which also confirms the higher standard deviation from the mean of prompt tuning performance. We can also see that for LLaMa-3.1-8B-Instruct and DeepSeek-LLM-7B-Chat, there is no statistically significant difference between using the original initialization or different task prompt vector initializations, and for SST2, CoLA, TREC, and MATH, average performance even slightly increased, but in most cases the performance remained unchanged, according to statistical significance tests. In some cases, the performance of the original initialization of the T5 model was similar or even better than for much larger instruction-fine-tuned LLaMa or DeepSeek models, which goes in line with findings of recent related work \cite{pecher2024comparing,gurgurov2025small}.

\begin{table}[t]
\setlength{\tabcolsep}{4pt}
\centering
\caption{Comparison of test results across 3 random soft prompt initializations for T5-base, LLaMa-3.1-8B, and DeepSeek-LLM-7B models. The first column (Original) represents the results of prompt tuning. The second column (Mixed) represents the results of moving a specific initialization in the direction of a task prompt vector created from different (mixed) initializations. N/A means that the task was too complex for the T5 model, and the results were underperforming.}
\resizebox{0.95\textwidth}{!}{%
\begin{tabular}{lc|ccc|ccc|ccc}
\toprule
\multicolumn{2}{c|}{\multirow{2}{*}{Task}}                           & \multicolumn{3}{c|}{T5}                   & \multicolumn{3}{c|}{LLaMa}               & \multicolumn{3}{c}{DeepSeek}           \\
\multicolumn{2}{c|}{}                                                & Original     & Mixed           & $\Delta$ & Original     & Mixed          & $\Delta$ & Original     & Mixed        & $\Delta$ \\ \midrule
\multirow{9}{*}{\rotatebox{90}{GLUE}}   & MNLI    & 85.4$_{0.1}$ & 85.3$_{0.2}$    & -0.1     & 89.7$_{0.2}$ & 89.7$_{0.2}$   & +0.0     & 86.1$_{1.9}$ & 86.0$_{2.0}$ & -0.1     \\
                                                           & QQP     & 87.3$_{0.1}$ & 87.4$_{0.1}$    & +0.1     & 84.6$_{0.1}$ & 84.6$_{0.1}$   & +0.0     & 84.4$_{0.1}$ & 84.4$_{0.1}$ & +0.0     \\
                                                           & QNLI    & 93.3$_{0}$   & 93.2$_{0.1}^*$  & -0.1     & 92.0$_{0}$   & 92.0$_{0.1}$   & +0.0     & 90.3$_{1.4}$ & 90.4$_{1.3}$ & +0.1     \\
                                                           & SST2    & 93.8$_{0.3}$ & 93.2$_{0.6}$    & -0.6     & 95.9$_{0.4}$ & 96.0$_{0.5}$   & +0.1     & 95.6$_{0.1}$ & 95.6$_{0.1}$ & +0.0     \\
                                                           & STS-B   & 89.3$_{0.2}$ & 88.6$_{0.2}^*$  & -0.7     & 89.9$_{1}$   & 89.8$_{0.8}$   & -0.1     & 88.7$_{0.3}$ & 88.7$_{0.4}$ & +0.0     \\
                                                           & MRPC    & 90.8$_{0.8}$ & 83.0$_{4.1}^*$  & -7.8     & 87.7$_{0.2}$ & 88.1$_{0.1}^*$ & +0.4     & 87.5$_{1.1}$ & 87.4$_{1.1}$ & -0.1     \\
                                                           & RTE     & 50.3$_{4.2}$ & 63.4$_{1.1}^*$  & +13.1    & 89.7$_{0.3}$ & 89.4$_{0.6}$   & -0.3     & 84.3$_{0.7}$ & 84.3$_{0.7}$ & +0.0     \\
                                                           & CoLA    & 85.9$_{0.2}$ & 84.9$_{0.3}^*$  & -1.0     & 87.3$_{1.6}$ & 87.6$_{1.2}$   & +0.3     & 87.3$_{0.6}$ & 87.6$_{0.5}$ & +0.3     \\
                                                           & avg     & 84.5$_{1}$   & 84.0$_{1.9}$    & -0.5     & 89.6$_{0.7}$ & 89.7$_{0.6}$   & +0.1     & 88.1$_{0.8}$ & 88.1$_{0.8}$ & +0.0     \\ \midrule
\multirow{5}{*}{\rotatebox{90}{Others}} & TREC    & 95.5$_{1.7}$ & 26.5$_{18.2}^*$ & -69.0    & 95.8$_{0.3}$ & 96.0$_{0.3}$   & +0.2     & 95.7$_{1.0}$ & 95.6$_{1.0}$ & -0.1     \\
                                                           & DBPedia & 99.1$_{0}$   & 99.0$_{0.1}^*$  & -0.1     & 99.2$_{0}$   & 99.2$_{0}$     & +0.0     & 99.1$_{0.1}$ & 99.1$_{0.1}$ & +0.0     \\
                                                           & Yelp    & 97.2$_{0}$   & 97.1$_{0.1}^*$  & -0.1     & 98.6$_{0.1}$ & 98.6$_{0.1}$   & +0.0     & 98.4$_{0.1}$ & 98.4$_{0.1}$ & +0.0     \\
                                                           & SQuADv2 & N/A          & N/A             & N/A      & 66.3$_{0.9}$ & 66.4$_{0.9}$   & +0.1     & 63.8$_{0.9}$ & 63.8$_{0.6}$ & +0.0     \\
                                                           & MATH    & N/A          & N/A             & N/A      & 36.8$_{0.2}$ & 36.9$_{0.1}$   & +0.1     & 32.1$_{0.1}$ & 32.2$_{0.1}$ & +0.1     \\ \bottomrule
\end{tabular}%
}
\label{tab:cross_origin}
\end{table}

\paragraph{Task prompts and task prompt vectors maintain good performance even if they do not always point to the exactly same location in the task subspace.} To see whether the trained task prompts end up in the same task sub-space, we evaluate cosine similarity across multiple random initializations. We train multiple task prompts for 3 different random initializations and each source task (60 task prompts in total), and compute the cosine similarity from trained task prompts for each combination of random initializations and for each combination of tasks. We then average this cosine similarity for each task combination across all random initialization combinations. If task prompts are initialized from different random initializations and point in different directions in the task sub-space, we should also witness this phenomenon with their corresponding task prompt vectors. Therefore, we repeat this process for task prompt vectors.

Results in Table \ref{tab:cross_origin} row 1 indicate that the downstream performance of prompt tuning on the source tasks across 3 different random initializations has a low standard deviation from the average. This shows that the task prompts end up in a subspace with sufficient task performance without necessarily residing in the same task subspace. In addition, we do not observe any difference in findings from experiments with NLI and NLG tasks. 

Subsequently, Figures \ref{fig:rq1_heatmap_tp} and \ref{fig:rq1_heatmap_tpv} show the comparison of cosine similarities between task prompts and task prompt vectors from different tasks, averaged over all random initialization combinations. The cosine similarities on the diagonal serve as a baseline for comparison with the cross-task cosine similarities. We can see that cosine similarities for both task prompts as well as task prompt vectors are higher for combinations of tasks that are from similar problem domains or have similar labels and data structures. Another observation is that the cosine similarity of task prompts vectors provides a better measurement (in comparison with task prompts) of actual tasks' similarity as well as of the performance that can be achieved by a transfer between them (see also Figure \ref{fig:rq2_combinations}). For example, QNLI and TREC exhibit a relatively high similarity for their task prompts and a low similarity for task prompt vectors, which appropriately reflects their mutual diversity. In addition, we notice in Figures \ref{fig:rq1_heatmap_tp} and \ref{fig:rq1_heatmap_tpv} that task prompt vectors generally achieve lower cosine similarities than task prompts. Based on our results, we cannot determine the reason for this difference, and it can be a potential subject of future research. 

More detailed and disaggregated cosine similarities of Figures \ref{fig:rq1_heatmap_tp} and \ref{fig:rq1_heatmap_tpv} can be found in Supplementary Material C, Figures 1, and 2. We also evaluated cosine similarities of task prompts and task prompt vectors for LLaMa-3.1-8B-Instruct and DeepSeek-LLM-7B-Chat in Supplementary Material C in Figures 3a, 3b, 4a, 4b.

\paragraph{Task prompt vectors from similar problems are more similar.} Additionally, we evaluate the similarity of different task prompt vectors across different tasks. Figure \ref{fig:rq1_heatmap_tpv} shows that certain pairs of tasks are more similar than others, reflecting the shared properties of these tasks, such as the same number of classes, the same labels, or solving a similar problem. Problem similarity can be seen in the MNLI--QNLI task prompt vectors, and a similarity in the number of classes is observed in the MNLI task prompt vector, which tends to have higher cosine similarity with task prompt vectors for tasks with more classes (e.g., DBPedia, TREC). Increased similarity can also be seen in tasks that have common data formats (e.g., question-based QQP and QNLI). We also notice that MNLI-QQP and QNLI-QQP have even higher similarity than some tasks from common problems (e.g., MNLI--QNLI). This shows that the similarity of task prompt vectors may also appear for more dissimilar tasks. However, this phenomenon only appears in the case of the T5 model, but not necessarily in the results for LLaMa and DeepSeek models (available in Supplementary Material C).

\subsection{Combination of Task Prompt Vectors via Addition for Multi-Task Transfer}
\label{sec:exp_combinations}
This section addresses the following research question: \textbf{RQ2:} \textbf{Can we combine multiple task prompt vectors and maintain multi-task performance on the source tasks?} 

To answer this research question, we investigate the prompt arithmetics by task prompt vector addition on 55 task pair combinations from the set of NLU datasets (\textit{MNLI, QQP, QNLI, SST2, STS-B, MRPC, RTE, CoLA, TREC Coarse, DBPedia, Yelp Polarity}). We also evaluate combinations of task prompt vectors in a simulated limited data environment by providing 0--100 training examples before evaluation on the test set.

\begin{figure}[tb]
    \begin{centering}
        \includegraphics[width=\textwidth]{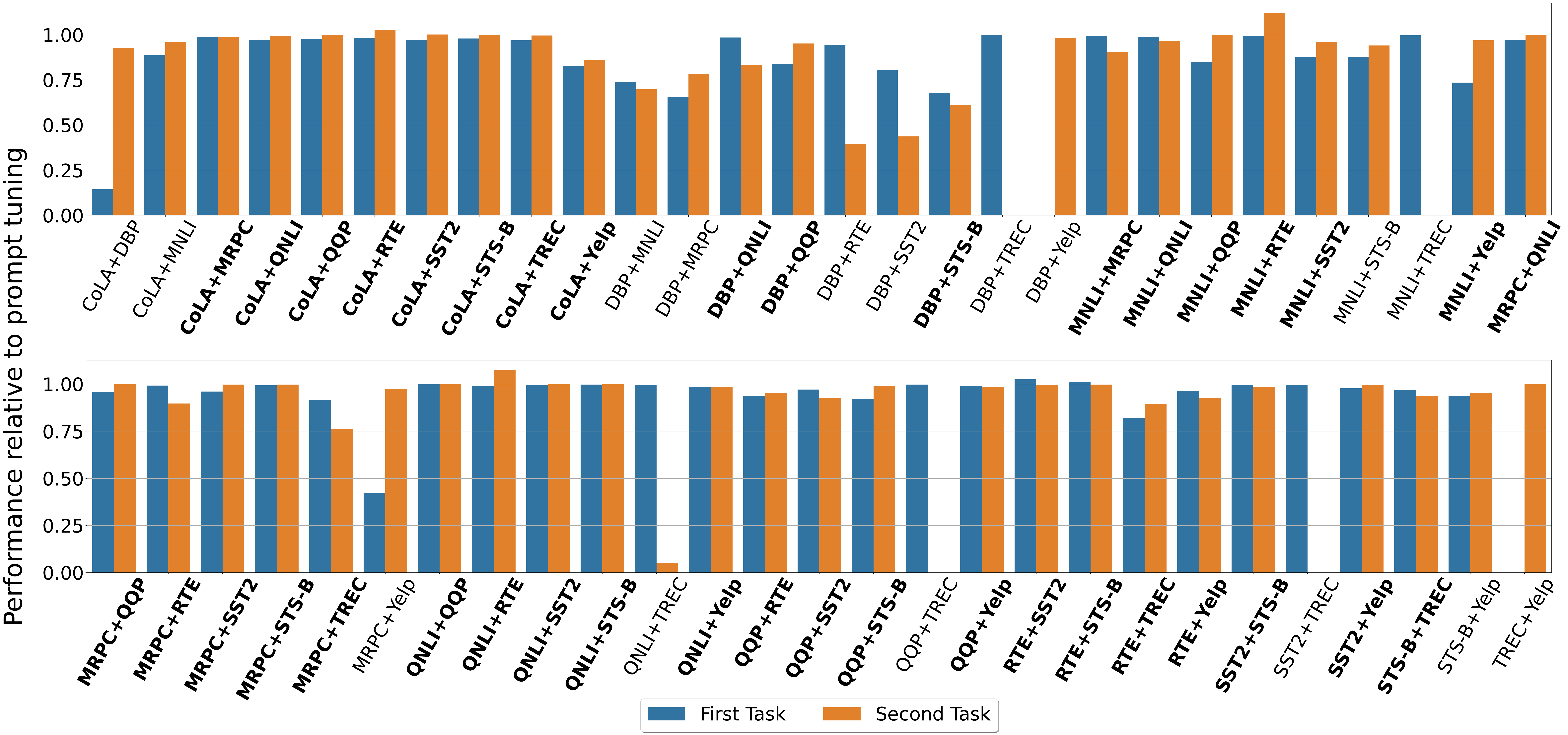}
        \caption{Comparison of relative exact match performance of combinations of task prompt vectors across averaged across 3 different random initializations and all task combinations. The results are relative to the original single-task performance (1 is the performance of single-task prompt tuning). The task combinations in bold are the combinations that achieved over 50\% of single-task performance on both of the tasks.\label{fig:rq2_combinations}}
    \end{centering}
\end{figure}

\paragraph{Combinations of task prompt vector pairs maintain good single-task performance on the majority of observed task combinations.} To evaluate whether combinations of task prompt vectors maintain single-task performance, we conduct experiments where we create paired combinations from all source tasks (according to equation \ref{eq:tpv_comb}). We can see from the results in Figure \ref{fig:rq2_combinations} that most binary classification tasks retain their single-task performance on both tasks, which implies that task prompt vectors can be used for solving multi-task problems, and also corresponds with previous finding that state that some tasks are mutually beneficial \cite{asai-etal-2022-attempt,vu-etal-2022-spot}. In some cases, the single-task performance was kept only for a single source task. This is, however, an expected behavior, because similar to other transfer learning approaches, a combination of task prompt vectors from too diverse tasks must inevitably end up in a negative transfer (e.g., for tasks with completely different features and meanings of labels). In some cases, the combination of two tasks even increased performance, for example, in the case of MNLI+RTE, possibly due to the shared task type (in this case, NLI/entailment). However, this increase is not clearly significant, as other NLI combinations do not show the same trend.

% Now that we know how the combinations of task prompt vectors affect the performance across source tasks, 
% we evaluate the combinations of task prompt vectors on a set of out-of-distribution target tasks from the same problem area.

\paragraph{Task prompt vector combinations are good initializations for zero-shot and few-shot learning.}% We conduct 0-shot and 100-shot evaluations of different initialization of prompt tuning on the set of target tasks. 
We select two target tasks for {inference classification} (\textit{SciTail, SNLI}), {topic classification} (\textit{AG News, Yahoo Answers}), and {sentiment classification} (\textit{SST5, IMDB}) while keeping the same set of source tasks. We compare initialization with randomly initialized soft prompts, soft prompts trained on single and multiple source tasks (equivalent to SPoT \cite{vu-etal-2022-spot}), the multi-task ATTEMPT \cite{asai-etal-2022-attempt} method, and a combination of task prompt vectors of both of the source tasks.

The 0-shot and 100-shot results (Table \ref{tab:zeroshots}) indicate that a combination of task prompt vectors can outperform initialization with a single-task source soft prompt on SciTail and IMDB, and the multi-task source soft prompt only for SciTail. The combination matches the SPoT baseline in cases like AG News, possibly because DBPedia and TREC together retain little TREC-specific information that could improve results. For SNLI, Yahoo Answers, and SST5 tasks, we can see that combinations of source task prompt vectors do almost match the results of the SPoT baseline.

\begin{table}[!t]
\centering
\caption{Test results of training T5-base model with random, single- and multi-task soft prompt transfer (SPoT), multi-task ATTEMPT, and our task prompt vectors on 0-shot and 100-shot data for all of our observed source and target tasks. We show the initialization with different combinations for NLI classification, topic classification, and sentiment classification. The subscript represents the standard deviation from the average. The best results are bold, while the second-best results are underlined. The * in the superscript represents that the results are statistically significant from the second-best result, by two-sample Student's t-test \cite{student1908probable}.}
\resizebox{0.9\textwidth}{!}{%
\begin{tabular}{lcc|lcc}
\toprule
\multicolumn{3}{c|}{SciTail (NLI)}                                                                      & \multicolumn{3}{c}{SNLI (NLI)}                                                                          \\ \midrule
\multicolumn{1}{c}{\multirow{2}{*}{Source tasks}} & \multicolumn{2}{c|}{F1}                             & \multicolumn{1}{c}{\multirow{2}{*}{Source tasks}} & \multicolumn{2}{c}{F1}                              \\
\multicolumn{1}{c}{}                              & 0 shots                  & 100 shots                & \multicolumn{1}{c}{}                              & 0 shots                  & 100 shots                \\ \midrule
Random                                            & $54.9_{6.6}$             & $75.6_{0.5}$             & Random                                            & $46.5_{1.5}$             & $47.6_{1.9}$             \\
MNLI (SPoT)                                       & $\underline{70.4_{0.4}}$ & $\underline{87.8_{0.9}}$ & MNLI (SPoT)                                       & $\underline{79.5_{0.3}}$ & $\underline{80.8_{0.4}}$ \\
QNLI (SPoT)                                       & $57.7_{13.1}$            & $77.7_{1.3}$             & QNLI (SPoT)                                       & $47.1_{0.3}$             & $49.1_{0.9}$             \\
QNLI + MNLI (SPoT)                                & $70.4_{1.2}$             & $87.7_{0.6}$             & QNLI + MNLI (SPoT)                                & $\mathbf{79.6_{0.2}}^*$  & $\mathbf{81_{0.4}}^*$    \\
QNLI + MNLI (ATTEMPT)                             & $63.8_{4.2}$             & $83.6_{3}$               & QNLI + MNLI (ATTEMPT)                             & $78.5_{0.5}$             & $79.6_{1.6}$             \\
QNLI + MNLI (ours)                                & $\mathbf{71.5_{0.8}}^*$  & $\mathbf{88.1_{0.9}}$    & QNLI + MNLI (ours)                                & $79.2_{1.4}$             & $80.3_{0.3}$             \\ \midrule
\multicolumn{3}{c|}{AG News (Topic)}                                                                    & \multicolumn{3}{c}{Yahoo Answers (Topic)}                                                               \\ \midrule
\multicolumn{1}{c}{\multirow{2}{*}{Source tasks}} & \multicolumn{2}{c|}{F1}                             & \multicolumn{1}{c}{\multirow{2}{*}{Source tasks}} & \multicolumn{2}{c}{F1}                              \\
\multicolumn{1}{c}{}                              & 0 shots                  & 100 shots                & \multicolumn{1}{c}{}                              & 0 shots                  & 100 shots                \\ \midrule
Random                                            & $0_{0}$                  & $50.4_{11.2}$            & Random                                            & $0_{0}$                  & $27.6_{10.6}$            \\
DBPedia (SPoT)                                    & $0_{0}$                  & $\mathbf{83.4_{0.6}}^*$  & DBPedia (SPoT)                                    & $0_{0}$                  & $\mathbf{61.3_{1.1}}^*$  \\
TREC (SPoT)                                       & $0_{0}$                  & $65.7_{5.6}$             & TREC (SPoT)                                       & $0_{0}$                  & $36.5_{8.7}$             \\
DBPedia + TREC (SPoT)                             & $0_{0}$                  & $82.1_{0.9}$             & DBPedia + TREC (SPoT)                             & $0_{0}$                  & $60.7_{2}$               \\
DBPedia + TREC (ATTEMPT)                          & $\mathbf{11.5_{1.7}}$    & $20.7_{2.8}$             & DBPedia + TREC (ATTEMPT)                          & $\mathbf{0.1_{0}}$       & $8.1_{5.6}$              \\
DBPedia + TREC (ours)                             & $0_{0}$                  & $\underline{83_{0.9}}$   & DBPedia + TREC (ours)                             & $0_{0}$                  & $\underline{61.1_{0.9}}$ \\ \midrule
\multicolumn{3}{c|}{IMDB (Sentiment)}                                                                   & \multicolumn{3}{c}{SST5 (Sentiment)}                                                                    \\ \midrule
\multicolumn{1}{c}{\multirow{2}{*}{Source tasks}} & \multicolumn{2}{c|}{F1}                             & \multicolumn{1}{c}{\multirow{2}{*}{Source tasks}} & \multicolumn{2}{c}{F1}                              \\
\multicolumn{1}{c}{}                              & 0 shots                  & 100 shots                & \multicolumn{1}{c}{}                              & 0 shots                  & 100 shots                \\ \midrule
Random                                            & $77.2_{9.6}$             & $89.4_{0.4}$             & Random                                            & $0_{0}$                  & $83.2_{5.8}$             \\
SST2 (SPoT)                                       & $88_{0.6}$               & $90.2_{0.3}$             & SST2 (SPoT)                                       & $\mathbf{94_{0.3}}^*$    & $\mathbf{93.9_{0.3}}^*$  \\
Yelp (SPoT)                                       & $90_{0.3}$               & $90.3_{0.2}$             & Yelp (SPoT)                                       & $88.6_{0.8}$             & $90.6_{0.5}$             \\
SST2 + Yelp (SPoT)                                & $\mathbf{90.8_{0.2}}$    & $\mathbf{90.8_{0.2}}$    & SST2 + Yelp (SPoT)                                & $\underline{93.7_{0.5}}$ & $\underline{93.8_{0.5}}$ \\
SST2 + Yelp (ATTEMPT)                             & $79.2_{6}$               & $89.4_{0.8}$             & SST2 + Yelp (ATTEMPT)                             & $16.4_{4.5}$             & $37.8_{7}$               \\
SST2 + Yelp (ours)                                & $\underline{90.1_{0.5}}$ & $\underline{90.4_{0.2}}$ & SST2 + Yelp (ours)                                & $89.9_{0.8}$             & $91.5_{0.5}$             \\ \bottomrule
\end{tabular}%
}
\label{tab:zeroshots}
\end{table}

ATTEMPT is also significantly underperforming when using a smaller set of pre-trained source soft prompts. Another observation is that ATTEMPT performs better on the AG News task. This may be caused by using the original implementation of ATTEMPT, where the authors, instead of using textual labels (i.e., "entailment", "not entailment"), used textual numbers as labels (i.e., "0", "1"), which makes the model predict numbers instead of specific words. %(unlike in the other methods).

\subsection{Additional Results: Few-Shot Comparison}
\label{sec:exp_ablation}
In this section, we study how increasing the number of demonstration data affects the performance of prompt tuning on a target task initialized by a combination of task prompt vectors of similar source tasks. We keep the same experiment setup as in the previous section and evaluate the soft prompt initialization on 5, 10, 25, 100, 250, and 500 shots.

\begin{figure}[!t]
    \begin{centering}
        \includegraphics[width=\textwidth]{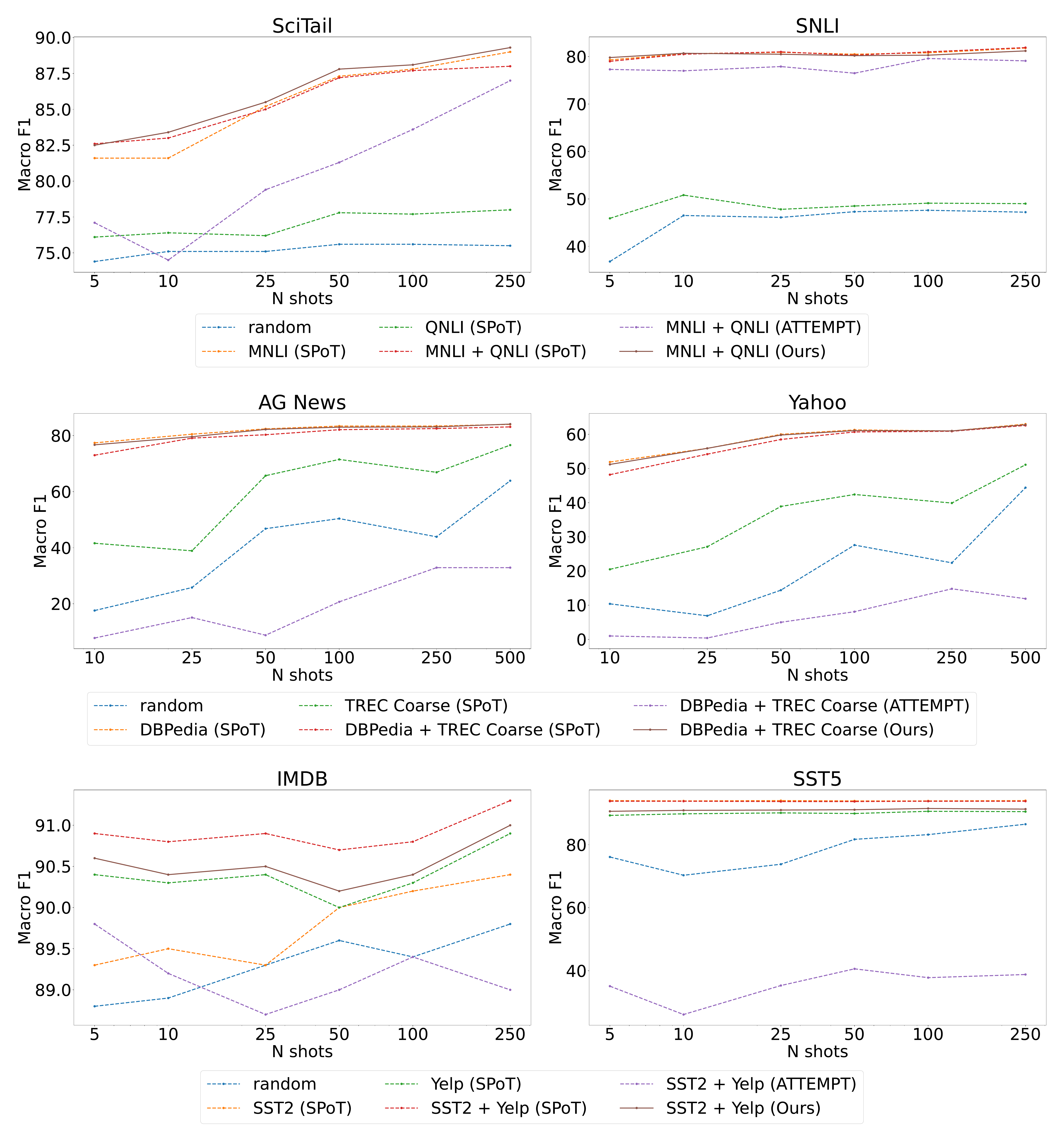}
        \caption{Test results of training T5-base model with random, single, and multi-task soft prompt transfer (SPoT), multi-task ATTEMPT, and our task prompt vectors combination on increasing numbers of shots of data. We can see that for SciTail and IMDB tasks, a combination of task prompt vectors outperforms single task transfer. \label{fig:fewshot}}
    \end{centering}
\end{figure}

The results in Figure \ref{fig:fewshot} indicate that the performance of the combination of task prompt vectors for SciTail and IMDB target tasks outperforms using a single-task initialization for multiple shots. We can also see that our method outperforms the multi-task initialization for the SciTail dataset across all shots of data.

Comparing the results from Figure \ref{fig:rq2_combinations} and Figure \ref{fig:fewshot}, if we choose a combination of tasks that maintains a significant amount of the source task performance (MNLI + QNLI and SST2 + Yelp), the few-shot performance of the task prompt vector combination tends to be higher than single-task transfer. In addition, we can see that in the case of the SST5 task, the SST2 initialization performs the best. We think that the reason for this may also be the similarity of SST5 and SST2, and that the combination of source tasks does not retain enough information to match the SST5 baseline.

\section{Discussion and Limitations}

\begin{table}[bt]
\setlength{\tabcolsep}{4pt}
\centering
\caption{Task prompt vectors maintain high task modularity and multi-task performance and are independent of the number of pre-trained source soft prompts.}
\resizebox{0.5\textwidth}{!}{%
\begin{tabular}{c|ccc}
\toprule
Method                                   & Modularity                            & \begin{tabular}[c]{@{}c@{}}Multi-task\\ performance\end{tabular}             & \begin{tabular}[c]{@{}c@{}}Source prompt\\ independence\end{tabular}            \\ \toprule
SPoT                                     & \Large\color{red}{\xmark} & \Large\color{green}{\cmark} & \Large\color{green}{\cmark}    \\
ATTEMPT                                  & \Large\color{green}{\cmark}    & \Large\color{green}{\cmark} & \Large\color{red}{\xmark} \\
\multicolumn{1}{l|}{\textbf{TPV (ours)}} & \Large\color{green}{\cmark}    & \Large\color{green}{\cmark} & \Large\color{green}{\cmark}    \\ \bottomrule
\end{tabular}%
}
\label{tab:comp_methods}
\end{table}

\paragraph{Comparison of task prompt vector properties with most relevant PEFT methods.} 
Table \ref{tab:comp_methods} compares attributes beneficial for multi-task training for SPoT, ATTEMPT, and task prompt vector methods. SPoT exhibits low multi-task modularity, with a need to re-train the source soft prompt every time the set of source tasks changes. ATTEMPT, while having sufficient task modularity, depends heavily on the quality and number of source soft prompts. While task prompt vectors, in general, are able to match the results of full multi-task soft prompt transfer (SPoT), initialization of prompt tuning using task prompt vector combinations also retains high task modularity, which means that new tasks can be added without the necessity of training, ultimately decreasing computational costs considerably. \textbf{Task prompt vectors thus have both -- modularity and source prompt independence -- and also retain sufficient multi-task performance.}

\paragraph{High reusability of the task prompt vectors.} 
In our experiments, we demonstrated multiple important properties of task prompt vectors. At first, we showed (Section \ref{sec:exp_create}) that task prompts and their corresponding task prompt vectors from different initializations do not necessarily point to the same space and that some vector combinations are more similar than others. Despite that, task prompt vectors created from one initialization and applied to a different initialization maintain their performance for the majority of observed tasks. The implication of this finding means that \textbf{it is possible to combine different task prompt vectors from different initializations}.

Furthermore, we showed (Section \ref{sec:exp_combinations}) that combinations of task prompt vectors for similar tasks maintain their source single-task performance (Figure \ref{fig:rq2_combinations}) and that the combinations of \textbf{task prompt vectors can be used for initialization of prompt tuning} in low resource settings (zero-/few-shot settings) on the set of target tasks (Table \ref{tab:zeroshots}). 

Based on both of these observations, we can very effectively re-use pre-trained task prompt vectors for different tasks and use them in downstream scenarios (even without a need for any further training). Since task prompt vectors are independent of their initialization, we can also \textbf{re-use pre-trained task prompt vectors shared by other researchers and practitioners} (e.g., on a designated vector hub).

\paragraph{Identification of appropriate source task prompt vectors.}
High reusability of task prompt vectors, however, requires identifying an appropriate source task prompt vector or a combination of them. To identify such a single vector/a combination of vectors, we propose to perform an evaluation on held-out validation sets. Another possible factor that can be included in the identification of an appropriate combination is the similarity of combined tasks. This similarity can be determined by data analysis by looking at commonalities in the task domain, data structure, or labels. Additionally, similarity can be quantified using the cosine similarities of task prompt vectors, which tend to correlate better with the resulting performance when compared to task prompts. 

\paragraph{Theoretical implications and analysis.} It lies beyond the scope of our work to further deliver theoretical analyses for diverse properties of task prompt vectors, which we will leave for future work. However, we still want to discuss some hypotheses that arise from the empirical results achieved during the experiments with task prompt vectors. 

At first, we can derive from the obtained findings that the \textbf{sub-space with optimal values in the soft prompt space has probably a convex shape}. This may be indicated by the fact that task prompts trained from different random initializations for the same task do not necessarily point in the same direction (based on Figures \ref{fig:rq1_heatmap_tp} and \ref{fig:rq1_heatmap_tpv}), but still achieve identical results.

Second, prompt arithmetics (task prompt vector addition) is possible even though the soft prompt space is non-linear. The rationale behind this could be that \textbf{task prompt vectors are linear approximations of how soft prompts change} during training. Another possibility may be that the task prompt vectors are sparse, and a combination of 2 sparse task prompt vectors creates a vector that contains more information about both tasks. These findings can be further useful for \textbf{machine unlearning tasks}, where one could also exploit task prompt vector subtraction.

\paragraph{Limitations.} To keep our focus on the evaluation of task prompt vectors, we utilize only monolingual models in the scope of our work, as well as 12 NLU and 2 NLG datasets in the English language only. Extension to multilingual models and datasets may reveal additional interesting findings about task prompt vectors features in multilingual settings.

In this work, we employed the set of 3 common NLU problems, each covering 4 different tasks, and 2 common NLG problems, covering 2 different tasks.
We consider this set as sufficient to evaluate the properties of task prompt vectors, also taking computational costs into account -- adding more tasks would also result in more computational costs. Nevertheless, additional tasks may still strengthen findings presented in this paper.

Even though there are many other PLMs capable of conditional generation that beat T5 models in performance on various benchmarks, we focus our experiments on the T5-base model as it is commonly used as a representative model in many PEFT methods. Additional experiments on a larger set of models, therefore, represent another potential extension of our work. 

\section{Conclusion}
In our work, we introduce and investigate task prompt vectors as a method of multi-task transfer from prompt tuning. We show that the task prompt vectors are not dependent on random initialization and that the performance across different random initializations does not change significantly in the majority of observed source tasks. We show that for similar and mutually related tasks, the combination via arithmetic addition maintains the single-task performance or even improves it. Finally, we show that certain combinations of task prompt vectors can be a better option for initialization for certain tasks while maintaining higher multi-task modularity than other soft prompt-based methods like SPoT and ATTEMPT.

In the future, we would like to evaluate the cross-model performance of task prompt vectors. We think that further experiments with generation tasks may be another interesting extension. Moreover, task prompt vector arithmetic has the highest potential for improving the unlearning in PLMs by negating the task prompt vectors for the tasks we want to unlearn. Such an option is enabled by introducing task prompt vectors, which would not be possible with the existing state-of-the-art methods.

\begin{credits}
\subsubsection{\ackname} This work was partially funded by European Union under the project DisAI, GA No. 101079164, and the project CEDMO 2.0, GA No. 101158609; by the European Union NextGenerationEU through the Recovery and Resilience Plan for Slovakia under the project No. 09I01-03-V04-00006; and by the Slovak Research and Development Agency under the Contract no. APVV-22-0414.

Part of the research results was obtained using the computational resources procured in the national project funded by the Ministry of Education, Youth and Sports of the Czech Republic through the e-INFRA CZ (ID:90254); and the national project National competence centre for high performance computing (project code: 311070AKF2) funded by ERDF, EU Structural Funds Informatization of Society, Operational Program Integrated Infrastructure.

The authors also wish to acknowledge the TAILOR project funded by the European Union under the EU Horizon 2020, GA No. 952215, which supported the research mobility that started the collaboration on this paper under the TAILOR Connectivity fund.
\end{credits}

% \bibliographystyle{splncs04}
% \bibliography{ref}

\end{document}